\newif\iftr 
\trtrue 

\documentclass{article}
\usepackage{spconf,amsmath,graphicx}
\usepackage{amsfonts} 
\usepackage{bm} 
\usepackage{enumitem} 
\usepackage{xcolor} 
\usepackage{cite}

\usepackage{amssymb,latexsym,amsthm} 
\usepackage{caption}
\usepackage{psfrag}
\usepackage{url}

\def\myrootdir{/Users/ayca/Dropbox/DRAFTS/}

\def\bibdirM{\myrootdir/MY_BIB}
\def\bibdirMM{\myrootdir/MY_BIB}

\newcommand{\newoperator}[3]{\newcommand*{#1}{\mathop{#2}#3}}

\newcommand{\tr}{\operatorname{tr}}
\newcommand{\E}{\mathbb E}
\newcommand{\prob}{\mathbb P}

\newoperator{\diag}{\mathrm{diag}}{\nolimits}
\newoperator{\rank}{\mathrm{rank}}{\nolimits}
\newoperator{\myperm}{\mathrm{perm}}{\nolimits}

\newoperator{\sgn}{\mathrm{sgn}}{\nolimits}

\newoperator{\abs}{\mathrm{abs}}{\nolimits}

\mathchardef\myhyphen="2D

\newcommand{\xbf}{\boldsymbol x}

\newcommand{\thetabf}{\boldsymbol \theta}

\newcommand{\Abf}{\boldsymbol A}
\newcommand{\Bbf}{\boldsymbol B}
\newcommand{\Ibf}{\boldsymbol I}

\newcommand{\zbf}{\boldsymbol z}
\newcommand{\ybf}{\boldsymbol y}

\newcommand{\Zbf}{\boldsymbol Z}

\newcommand{\Hbf}{\boldsymbol H}

\newcommand{\Fbf}{\boldsymbol F}

\newcommand{\Wbf}{\boldsymbol W}

\newcommand{\Vbf}{\boldsymbol V}

\newcommand{\omegabf}{\boldsymbol \omega}

\newcommand{\Kbf}{\boldsymbol K}

\newcommand{\zerobf }{\boldsymbol 0}

\newcommand{\herm}{\mathrm{H}}
\newcommand{\trans}{\mathrm{T}}

\newcommand{\pinv}{\mathrm{\dagger}}

\newcommand{\Dset }{\mathcal{D}}

\newcommand{\Nset }{\mathcal{N}}

\newcommand{\Cbb }{\mathbb{C}}
\newcommand{\Rbb }{\mathbb{R}}

\newoperator{\myvec}{\mathrm{vec}}{\nolimits}

\newcommand{\zcw}{ \bar{\zbf} } 
\newcommand{\Einj}{\mathcal{E}_I}
\newcommand{\Enorm}{\mathcal{E}_P}
\newcommand{\EBP}{\mathcal{E}_{BP}}
\newcommand{\Einjnorm}{\mathcal{E}_{IP}}

\definecolor{myblue}{rgb}{0.3328,    0.3539,    0.7758}
\definecolor{myblue2}{rgb}{0.0328,    0.0539,    0.4758}
\definecolor{mygreen2}{rgb}{ 0.0328 0.4758    0.0539} 
\definecolor{mygreen3}{rgb}{ 0.0328 0.1758    0.0539} 
 \definecolor{myred}{rgb}{0.4758, 0.0328,    0.0539}

\newtheorem{thm}{\bf{Theorem}}[section]

\newtheorem{lem}{\bf{Lemma}}[section]

\theoremstyle{remark} 

\newtheorem{rem}{\bf{Remark}}[section]
\newtheorem{ex}{\bf{Example}}[section]

\newenvironment{theorem}
{\vspace{3pt}\par\noindent \thm \begin{itshape}\noindent}
{\end{itshape} \vspace{6pt}}
\newenvironment{lemma}
{ \vspace{3pt} \par\noindent  \lem \begin{itshape}\noindent}
{\end{itshape} \vspace{6pt}}
\newenvironment{remark}
{\vspace{2pt} \par\noindent \rem} 
{\vspace{2pt}}

\newenvironment{example}
{\vspace{3pt} \par\noindent \ex} 
{\vspace{6pt}}

\title{Sparse Recovery with Non-linear Fourier Features}

\name{Ay\c ca \"Oz\c celikkale \thanks{A.~\"Oz\c celikkale acknowledges the support from Swedish Research Council under grant 2015-04011.} }
\address{Signals and Systems, Uppsala University, Sweden}

\begin{document}
%
\maketitle
\begin{abstract}
Random non-linear Fourier features have recently shown remarkable performance  in a wide-range of regression and classification applications. Motivated by this success,  this article focuses on a sparse non-linear Fourier feature (NFF) model.  We provide a characterization of  the sufficient number of data points  that guarantee perfect  recovery of the unknown parameters with high-probability.  In particular, we show how the sufficient number of data points depends on the  kernel matrix associated with the probability distribution function of the input data. We compare our results with the recoverability bounds for the bounded orthonormal systems and provide examples that illustrate sparse recovery under the NFF model. 
\end{abstract}

\begin{keywords}
Random Fourier features, kernels, sparsity, compressive sensing.
\end{keywords}

\kern-1em
\section{Introduction}
\kern-0.4em
\label{sec:intro}

In the canonical statistical learning problem, we have access to  $(\xbf_i,y_i)$ pairs where we have statistically  independent and identically distributed (i.i.d) $\xbf_i$ and the corresponding  $y_i$, $i= 1,\ldots, M$.  Here, $\xbf_i $ denotes the known input, and $y_i$ denotes the associated responses/labels.  The standard aim of the learning problem is to construct a function $f(\xbf)$ to predict the relevant response $y$ given a previously unobserved input $\xbf$.  

In this article,  we consider the above learning problem when the data pairs $(\xbf_i,y_i)$ come from a sparse non-linear Fourier feature (NFF) model.
NFFs have been proposed by \cite{RahimiRecht_2008} to provide efficient approximations of the kernel methods. Algorithms that utilize NFFs have shown remarkable performance in regression and classification applications in a wide range of real-world data scenarios including MNIST image data,   census data, network intrusion detection and human activity recognition \cite{RahimiRecht_2008,RahimiRecht_2008randomkitchenSinks,BelkinHsuMaMandal_2019}. The success of NFFs in these applications suggests that the random NFF model provides a suitable model for real-world data.

Motivated by this success,  we investigate the conditions  that lead to perfect  recovery of unknown parameters  when the data comes from a sparse NFF model.   
%
In our main result, we provide a characterization of  the sufficient number of data points  that guarantee perfect  recovery of the unknown parameters with high-probability.
In particular, we show how the sufficient number of data points depend on the  kernel matrix associated with the probability distribution function of the input data. 

Overview of  the related work and the contributions: Performance of NFFs have been investigated in a number of recent works.
Guarantees  for kernel approximation \cite{RahimiRecht_2008}  and statistical guarantees for kernel ridge regression \cite{RahimiRecht_2008randomkitchenSinks,AvronKapralovMuscoMuscoVelingkerZandieh_2018,AlaouiMahoney_2015} 
have been provided.  Connections between the NFFs and Gaussian processes have been explored \cite{HensmanDurrandeSolin_2017}. 
Spectral properties of general random nonlinear transformations have been investigated \cite{LiaoCouillet_2018spectrum}.  
Behaviour of the NFF-based solutions under norm constraints has been the attention of a number of recent works  \cite{BelkinHsuMaMandal_2019,belkinHsuXu_2019}.  Here,  we contribute to this last line of work by considering methods that  directly minimize the $l_1$-norm of the unknown parameters  and by providing sufficient conditions for recovery with high probability. 
We also compare our results with the performance guarantees for the bounded orthonormal systems  \cite{foucartRauhut_2013,CandesRomberg_2007} and provide examples that illustrate sparse recovery under the NFF model. 
%

Notation: We denote a column vector  of size  $N \times 1$ with $\mathbf{a}=  [a_1; \ldots;  a_N] \in \Cbb^{N \times 1}$ where semi-colon $;$  is used to separate the rows.  
Complex conjugate transpose, the tranpose, and  the pseudo-inverse of a matrix $\Abf$  is denoted by $\Abf^\herm$,  $\Abf^\trans$ and $\Abf^\pinv$,  respectively. Spectral norm of a matrix is denoted by $||\Abf||$. The $l^{th}$ row, $k^{th}$ column element of a matrix $\Abf$ is denoted by $A_{lk}$. The $N \times N$ identity matrix is denoted by $\Ibf_N$.  The largest and the smallest eigenvalues are denoted by $\lambda_{max}(\Abf)$ and $\lambda_{min}(\Abf)$, respectively.   
%

\kern-0.6em
\section{Signal Model and Problem Statement}
\kern-0.6em

Consider the statistical learning problem described in Section~\ref{sec:intro}. We assume that the data comes from a  non-linear Fourier features model. 
In particular, let $\Omega =\{ \omegabf_1, \ldots,  \omegabf_N \}$  denote the set of frequencies where $\omegabf_k \in \Rbb^{d \times 1}$ is the $d$-dimensional frequency variable. 
The relationship between the input $\xbf \in \Rbb^{d \times 1}$ and the output $y \in \Rbb$ is given as 
\begin{align}\label{eqn:truemodel}
y= f_{\bar{\thetabf}}(\xbf) = \sum_{k=1}^N \bar{\theta}_k \phi(\xbf, \omegabf_k)=  \frac{1}{\sqrt{N}} \sum_{k=1}^N \bar{\theta}_k   e^{-j  \omegabf_k^T \xbf},
\end{align}
where $ \phi(\xbf, \omegabf_k) = \frac{1}{\sqrt{N}}e^{-j \omegabf_k^T \xbf}$, $j\!=\!\sqrt{-1}$ denotes the Fourier features. Here,   $\bar{\thetabf}= [\bar{\theta}_1; \ldots; \bar{\theta}_N]  \in \Cbb^{N \times 1}$ denotes the true model parameters and  $f_{\bar{\thetabf}}(.)$ denotes the associated true data model function.
We assume that $\bar{\thetabf}$ is $D$-sparse, i.e.  at most $D$ of $\bar{\theta}_i$'s  are possibly non-zero. 

We have access to $M$ input-output pairs $ (\xbf_i, y_i)$ with 
\begin{align}
y_i = f_{\bar{\thetabf}}(\xbf_i), \quad i=1,\ldots,M,
\end{align}
where   $\xbf_i$'s are i.i.d. with $\xbf \sim p(\xbf)$.
We would like to recover the unknown model parameters $\bar{\thetabf}$ using this data. 
To approximate  $f_{\bar{\thetabf}}(\xbf)$, we use $f_{{\thetabf}}(\xbf)$ defined as follows
\begin{align}\label{eqn:f}
 f_{{\thetabf}}(\xbf) = \sum_{k=1}^N \theta_k \phi(\xbf, \omegabf_k), 
 \end{align}
where $\thetabf = [\theta_1; \ldots; \theta_N]  \in \Cbb^{N \times 1}$ denotes the coefficients that we optimize over to fit to the data.

To find $\bar{\thetabf}$, we focus on the following basis pursuit formulation
\begin{subequations}\label{eqn:bp} 
\begin{align}
 \,\, \min_{\,\, \substack{\thetabf \in \Cbb^{N \times 1}}}  \quad & || \thetabf ||_1  \\
\text{s.t.} \quad
\label{eqn:cons}
&  f_{\bar{\thetabf}}(\xbf_i)  = f_{{\thetabf}}(\xbf_i), \quad \quad  i= 1,\ldots,M.
\end{align}
\end{subequations}
Note that in \eqref{eqn:cons},  $f_{\bar{\thetabf}}(\xbf_i)\!=\!y_i$ denotes the observations/data and $f_{{\thetabf}}(\xbf_i)\!=\! \sum_{k=1}^N \theta_k \phi(\xbf_i, \omegabf_k) $ denotes the fitted model whose coefficients we optimize over. 

The observations  $\ybf =[ y_1; \ldots; y_M ]  \in \Cbb^{M \times 1}$  can be expressed as $ \ybf =\Zbf \bar{\thetabf}$, where    the elements of $\Zbf \in \Cbb^{M \times N}$ are given by $Z_{i,k} = \frac{1}{\sqrt{N}} e^{-j \xbf_i^T\omegabf_k}$. 
Hence, the basis pursuit formulation in \eqref{eqn:bp} can be equivalently expressed in terms of  $\Zbf$ as follows: 
\begin{subequations}\label{eqn:bp:z} 
\begin{align}
 \,\, \min_{\,\, \substack{\thetabf \in \Cbb^{N \times 1}}}  \quad & || \thetabf ||_1  \\
\text{s.t.} \quad
\label{eqn:cons:z}
&  \Zbf \bar{\thetabf}  = \Zbf \thetabf.   
\end{align}
\end{subequations}
Here  $\Zbf \bar{\thetabf} $ denotes the observations ($f_{\bar{\thetabf}}(\xbf_i),\, i=1, \ldots, M$) and  $\Zbf \thetabf $ denotes the model whose coefficients we optimize over ($f_{{\thetabf}}(\xbf_i),\, \, i=1, \ldots, M$).

We provide our main result, i.e. statistical performance guarantees for this basis pursuit formulation, in Thm.~\ref{thm:hp:general}. 
Besides basis pursuit, our  analysis also holds  for other popular sparsity inducing algorithms, see Remark~\ref{remark:alg}.

\kern-1em
\section{Performance Guarantees}

\subsection{Preliminaries: Fourier Features and Kernel Matrices}
%
Using Bochner's theorem \cite{RahimiRecht_2008},  we consider the real-valued shift-invariant kernel  $k(\omegabf,\tilde{\omegabf})$  on $\Rbb^d  \times \Rbb^d$ associated with the symmetric probability distribution $p(\xbf)$  as follows
\begin{align}\label{eqn:kernel:FT}
k(\omegabf, \tilde{\omegabf}) =  \int_{\Rbb^d} p(\xbf) e^{-j \xbf^T(\omegabf-\tilde{\omegabf})} d\xbf.
\end{align}
Let us denote  the shift-invariant kernel $k(\omegabf, \tilde{\omegabf})$ with $k(\omegabf, \tilde{\omegabf})= k(\omegabf- \tilde{\omegabf}) = k (\Delta \omegabf)$ where  $\Delta \omegabf=\omegabf- \tilde{\omegabf}$.
Hence, \eqref{eqn:kernel:FT} states that $k (\Delta \omegabf)$  is the Fourier transform of  $p(\xbf)$, and equivalently $k (\Delta \omegabf)$ is the characteristic function of $\xbf$. 
For instance, for $\xbf$ Gaussian with  $\xbf \sim \mathcal{N}(0, \sigma^2 \Ibf_d)$,  we have 
\begin{align}\label{eqn:Gaussian:k}
k(\omegabf,\tilde{\omegabf})= e^{- \frac{\sigma^2}{2}||\omegabf-\tilde{\omegabf}||_2^2},
\end{align}
that is, the squared exponential (i.e. Gaussian) kernel \cite{RahimiRecht_2008}. 
Similarly, exponential kernel and the Cauchy kernel can be constructed from Caucy distribution and the Laplace distribution, respectively \cite{RahimiRecht_2008}. 
Here,  \eqref{eqn:kernel:FT} always speficies a normalized kernel, i.e. $k(\omegabf, \omegabf) = k (\zerobf) =1 $, since with $\omegabf= \tilde{\omegabf}$, \eqref{eqn:kernel:FT} becomes the integral over the probability distribution $p(\xbf)$ over $\Rbb^d$.
%

We note that \eqref{eqn:kernel:FT} can be expressed as
\begin{align}
\label{eqn:expectation}
k(\omegabf,\tilde{\omegabf}) &=\E[e^{-j \xbf^T(\omegabf-\tilde{\omegabf})} ].
\end{align}
Here the expectation is over random data $\xbf$. 
Note that  our point of view is different from  \cite{RahimiRecht_2008} where expectation over randomly chosen $\omegabf$'s is used to provide approximations of the kernel for a given set of data.

The kernel matrix associated with $\Omega$, i.e.  $\Kbf \in \Rbb^{N \times N}$ has the elements
\begin{align}
K_{ij}=k(\omegabf_i- \omegabf_j) = \E[e^{-j \xbf^T(\omegabf_i-\omegabf_j)} ].
\end{align}
Note that $\Kbf \succeq 0$. In the below, we assume that  $\Omega$ has distinct frequencies  ( $\omega_i = \omega_l \Leftrightarrow i=l $) and $\Kbf \succ 0$.  
Note that we have $K_{ij}  \leq 1,\, \forall i,j$, where the on-diagonal elements are given by   $K_{ii}= k(\omegabf_i, \omegabf_i) =1$.  
 Let $k_{max}$ be the largest off-diagonal element in absolute value, i..e  $k_{max} = \max_{i,j;\\ i\neq j} |K_{ij}|$. 
We denote the condition number of $\Kbf$ with   
\begin{align}
\beta \triangleq \frac{\lambda_{max}(\Kbf) } {\lambda_{min}(\Kbf)}.
\end{align}

\subsection{Main Result: Recovery with High Probability}

We now present some notation for our main result. 
The set of indices for which $\bar{\theta}_i$ is possibly non-zero is denoted by $\Dset$. The corresponding frequency subset is denoted by $\Omega_D$, $\Omega_D \subseteq \Omega$. We denote the vector with $D$ elements which only consists of the coefficients whose indices are in  $\Dset$ with $\bar{\thetabf}_D$.   Let $\sgn(\bar{\thetabf})$ denote the vector of signs of the elements $\bar{\theta}_i$, where the sign is defined as $\bar{\theta}_i/ |\bar{\theta}_i|$ if $\bar{\theta}_i \neq 0$, and as $0$ otherwise. 
%

Our main result is the following: 

\begin{theorem}\label{thm:hp:general}
 Assume that we have access to i.i.d. $\xbf_i$  data with $\xbf_i \sim p(\xbf)  $ and the corresponding responses  $y_i$, $i=1, \ldots, M$ from the model \eqref{eqn:truemodel}.  Given $\Omega$ with distinct frequencies,  let $\Kbf \succ 0$ with $K_{i,j } = k(\omegabf_i, \omegabf_j) $ be  the associated kernel matrix. 
 Let $2 D \leq N$.
Let $\bar{\thetabf} \in \Cbb^{N \times 1}$ denote the D-sparse vector of unknown coefficients in \eqref{eqn:truemodel} with  such that $\sgn(\bar{\thetabf}_D)$ forms a Rademacher or Steinhaus sequence.  Let the number of data points $M$ satisfy  $M \geq M_k$ where
\begin{align}\label{eqn:M}
M_k \triangleq   C \times D \times  \ln(3 {N / \delta}),
\end{align}
 $C =C_{q} \times C_{\beta}$, 
$C_{q} = (\frac{1+q C_{\eta}}{\lambda_{min}(\Kbf) - q  \sqrt{D} k_{max}})^2 $,  
$ C_{\beta} =2  (\beta+\frac{2}{3}) \lambda_{min}(\Kbf)$, 
$C_{\eta} = \sqrt{\frac{28}{3} \frac{1}{C_\beta}}$, 
$q=\sqrt{2 \ln (6 N/\delta)}$, 
$ C_{\eta}/\sqrt{C_{q}}  \leq 2 \sqrt{D} $, $\lambda_{min}(\Kbf) \geq q  \sqrt{D} k_{max}$.
Then, with probability at least $1-\delta$, the unique minimizer of \eqref{eqn:bp} gives the true parameter vector $\bar{\thetabf}$. 
\end{theorem}

The proof is presented in Section~\ref{pf:thm:hp:general}. The constant $C$ only depends on the properties of the whole frequency set $\Omega$ but not  on the unknown subset $\Dset$. Hence, the conditions of the theorem can be evaluated using only $\Kbf$. Note that $\Kbf$ depends on $p(\xbf)$ but not on the realizations of $\xbf$.

In Thm.~\ref{thm:hp:general}, a crucial point is how large $C$ is.  In Section~\ref{sec:DFT}, we investigate this point by  comparing $M_k$ with the results for the well-established bounded orthonormal system scenario, in particular with the Discrete Fourier transform (DFT) case \cite[Ch.~12]{foucartRauhut_2013}, \cite[Thm.~1.1]{CandesRomberg_2007}.

\begin{remark}\label{thm:hp:general}
 Consider the true risk, i.e. $\E[ l(f_{{\thetabf}}(\xbf), y)]$ where $l(f_{{\thetabf}}(\xbf), y)$ is a cost function, such as the quadratic cost $(f_{{\thetabf}}(\xbf)- y)^2$.   Thm~\ref{thm:hp:general} shows that (under the given conditions), true risk  is zero with high probability since the true coefficient vector $\bar{\thetabf}$ can be recovered perfectly.  
\end{remark}

\begin{remark}\label{remark:alg}
By \cite[Prop. 3.2]{foucartRauhut_2013}, Thm~\ref{thm:hp:general} guarantees that   there exist appropriate parameters for the other popular compressive sensing algorithms (basis pursuit denoising, quadratically constrained denoising and least absolute shrinkage and selection operator (LASSO) ) so that the true coefficient vector $\bar{\thetabf}$ is recovered with probability $1-\delta$. 
\end{remark}

\begin{remark}
In general, $\Kbf$ is different from the identity matrix. This distinguishes the scenario here from the case of bounded orthonormal systems \cite[Ch.12]{foucartRauhut_2013}.  Nevertheless, $M_k$  depends on how close $\Kbf$ is to the identity matrix through the condition number $\beta$, the minimum eigenvalue $\lambda_{min}(\Kbf)$ and the magnitude of the off-diagonal elements $k_{max}$.  
\end{remark}

\subsection{Comparison with the Recovery Guarantees under Randomly Sampled DFT}\label{sec:DFT}
We now compare the condition on the number of data points in Thm.~\ref{thm:hp:general} with the recovery results for the bounded orthonormal systems. In particular,  we consider the case of the DFT with $d=1$. Let $ \Fbf \in \Cbb^{N \times N}$  be the DFT matrix, i.e. 
\begin{align}
\Fbf_{tk} ={1 \over \sqrt{N}} e^{-j \frac{2 \pi}{N} (t \myhyphen 1) (k \myhyphen 1)} \quad 1 \leq t,k \leq N.
\end{align} 
The observations are given by \cite[Ch.12]{foucartRauhut_2013},\cite{CandesRomberg_2007}
\begin{align} \label{eqn:DFTmodel}
 \ybf_f= \Hbf  \Fbf \bar{\thetabf},
 \end{align}
where  $\Hbf \in \Rbb^{M \times N}$ is a random sampling matrix, i.e. a rectangular diagonal matrix.  We have $H_{li}=1$ if and if  $i^{th}$ component of  $ \Fbf \bar{\thetabf} \in \Cbb^{N \times 1} $ is measured in the $l^{th}$ measurement.   Hence, the problem is to reconstruct the vector $\bar{\thetabf}$  from   $\ybf_f$, i.e. from $M$ randomly selected entries of its discrete Fourier transform $\Fbf \bar{\thetabf} $.  Note that  the DFT is one of the unitary transforms that has the smallest coherence,  and hence yields the most favorable sufficient conditions  for recoverability among the discrete unitary transforms  \cite[Ch.12]{foucartRauhut_2013},\cite{CandesRomberg_2007}. 

The scenario in \eqref{eqn:DFTmodel}  can be also interpreted as a special case of a discrete counterpart of the NFF scenario in \eqref{eqn:truemodel} where $d=1$,  $p(x)$ is defined over $1, \ldots, N$ instead of over $\Rbb$,  and $\omega_i$ are equally spaced over $2 \pi$.

\begin{figure}
\begin{center}
\includegraphics[width=0.85\linewidth]{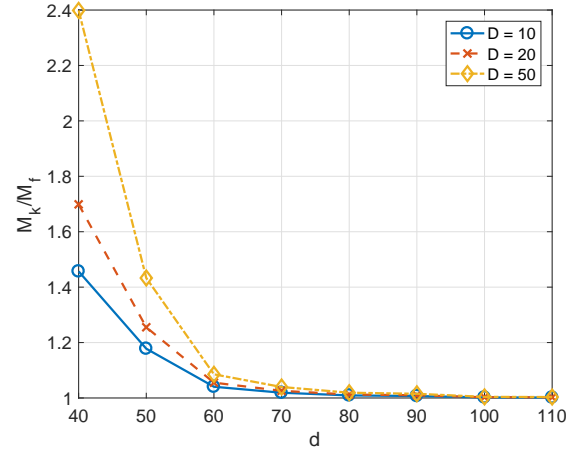}
\end{center}
\caption{ The ratio $M_k \over M_f$ versus $d$ with varying sparsity levels $D$
}
\label{fig:Mcomparison}
%
\end{figure}

Using basis pursuit, $\bar{\thetabf}$ can be recovered from  $\ybf$  with probability at least $1-\delta$ if  $M$ satisfies  $M \geq M_f  $ where
\begin{align}\label{eqn:M:DFT}
M_f= C' \times D  \times  \ln^2(6 {N / \delta}),
\end{align}
and $C' \leq 35$  \cite[Thm.~12.11]{foucartRauhut_2013}.   The next example compares $M_k$ with $M_f$.

\begin{example}
Consider the Gaussian kernel in \eqref{eqn:Gaussian:k}. Let $N=10^3$, $\sigma^2=1$,  $\delta=0.1$. For $M_k$, we randomly generate $\Omega$  and keep it fixed during the experiment.  We present  $M_k/M_f$ versus $d$ curves in Fig.~\ref{fig:Mcomparison}. 
For a more fair comparison, we also re-evaluate the analysis of $M_f$ on \cite[pg.388]{foucartRauhut_2013} which yields to a smaller $M_f$. 
Since the DFT case is a well-known scenario with good recoverability properties, $M_k/M_f \approx 1$ suggests that $\eqref{eqn:cons:z}$ provides a suitable data acquisition model for sparse recovery. In contrast,  large $M_k/M_f$ ratios indicate that higher number of measurements compared to the DFT case are needed with NFFs.  In Fig.~\ref{fig:Mcomparison}, we observe that  as $d$ increases, $M_k$ gets closer to $M_f$. For $d \gtrsim 80$, we have $M_k \approx M_f$.  This behaviour with increasing $d$ is consistent with the fact that as $d$ increases, $\Kbf$ becomes closer to $\Ibf_N$ (for fixed $\sigma^2$), see also Example~\ref{example:Gauss:highvariance}. We note that typical values of $d$ can be quite high, for instance applications using the popular benchmark case of image classification on MNIST database typically uses  $d=784$, e.g. \cite{BelkinHsuMaMandal_2019}.  
\end{example}

We now compare the following limiting case for the NFF setting with the DFT scenario:
\begin{example}\label{example:Gauss:highvariance}
Let $\xbf$ be Gaussian with  $\xbf \sim \mathcal{N}(0, \sigma^2 \Ibf_d)$, and hence we have the kernel in \eqref{eqn:Gaussian:k}. Let $\Omega$ consist of distinct frequencies. Consider the case with $ \sigma^2 \to \infty$, which yields to $K_{il}=k(\omegabf_i, \omegabf_l) \approx 0$ for $i \neq l$. Hence, the kernel matrix becomes $\Kbf \approx \hat{\Kbf} = \Ibf_N $. Using $\hat{\Kbf}$ instead of $\Kbf$ in   Thm~\ref{thm:hp:general}, we obtain the sufficient number of data points in \eqref{eqn:M} as
\begin{align}
M_k^g  \approx  C_g \times D \times  \ln(3 {N / \delta}),
\end{align}
where $C_g= {10 \over 3} \times (1+\sqrt{14\over 5} \sqrt{2 \ln (6 N/\delta)} )^2$. 
By straightforward algebraic manipulations, it can be shown that the condition $M \geq  M_k^g$ is  implied by 
\begin{align}\label{eqn:M:G}
M \geq   C_g' \times D \times  \ln^2(6 {N / \delta}),
\end{align}
where $C_g' \leq 30$. This condition is the same with the condition for recovery from partial Fourier measurements  in  \eqref{eqn:M:DFT} \cite[Thm.~12.11]{foucartRauhut_2013}. (The slightly tighter constant $C_g'$ in \eqref{eqn:M:G} is only due to the method of trivial algebraic manipulations, the same constant can be derived also for \eqref{eqn:M:DFT}).   Hence, as   $ \sigma^2 \to \infty$ (and hence as $\Kbf  \to  \Ibf_N$),   the behaviour of the model with NFFs becomes close to a partial Fourier measurement system in \eqref{eqn:DFTmodel}, and the sufficient number of measurements given in Thm~\ref{thm:hp:general}  becomes the same with the sufficient number measurements for recovery from partial Fourier measurements.
\end{example}

\kern-1em
\section{Numerical Results}\label{sec:num}
\kern-0.6em
We now illustrate recovery of $\bar{\thetabf}$ using \eqref{eqn:bp} under the NFF model.  Let $\xbf$ be Gaussian i.i.d. with  $\xbf \sim \mathcal{N}(0, \sigma^2 \Ibf_d)$. Hence, we consider the Gaussian kernel in \eqref{eqn:Gaussian:k}.
Let $N=500$, $\sigma^2=1$. We randomly generate $\Omega$ as i.i.d. multi-variate zero-mean Gaussian with uncorrelated components with variance $1$ and keep it fixed during the experiment. The $D$ non-zero elements of  $\bar{\theta}$ are generated i.i.d. from the uniform distribution  $\mathcal{U}[0,1]$. The $D$ locations of the possibly non-zero elements of  $\bar{\thetabf}$ are chosen randomly. The square-error is calculated as $\sum_{i=1}^N  ( \bar{\theta}_i -\theta_i )^2$.  At each simulation, a new set of data $(\xbf_i, y_i)$, $i=1, \ldots,M$  is formed and $\eqref{eqn:bp}$  is solved \cite{SeDuMi,cvx}. We perform $50$ Monte Carlo simulations and report the averages, i.e., the mean square-error (MSE).

\begin{figure}
\begin{center}
\psfrag{DATADATA1}{\tiny $M=100$}
\psfrag{DATADATA2}{\tiny $M=200$}
\includegraphics[width=0.9\linewidth]{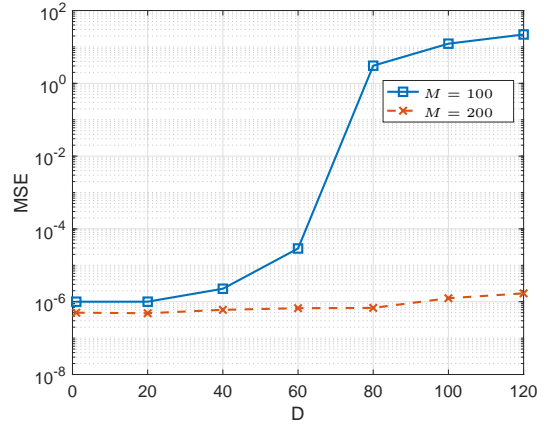}
\end{center}
\caption{ The MSE versus sparsity level $D$
}
\label{fig:MSE_d40}
%
\end{figure}

In Fig.~\ref{fig:MSE_d40}, we present the plots for $d=20$. When the number of data points is $M=100$, the MSE starts to increase after $D \approx 60$, indicating more measurements are needed for perfect recovery.  On the other hand,  with $M=200$,  low values of MSE are obtained for all $D$ values on the plot including the case with $D=120$, where the level of sparsity is low, i..e the data has relatively high degrees of freedom.  

\kern-0.5em
\section{Conclusions}
\kern-0.5em
Under a sparse non-linear Fourier features model, we have presented  bounds on the sufficient number of data points for recovery of the unknown coefficients with high probability. 
We have compared our bounds with the well-established case of bounded orthonormal systems. We have illustrated how the gap between the number of sufficient data points for the NFF scenario and the DFT scenario depends on the signal model parameters.

\kern-1em
\section{Appendix}\label{pf:thm:hp:general}
\kern-0.5em
\iftr
We first provide an overview.  Further details, including the proofs of Thm.~\ref{thm:invertible} and Thm.~\ref{thm:normsmall}, are provided in Section~\ref{sec:pf:thm:invertible} - Section~\ref{sec:pf:combination}.   %
\fi

We denote the $k$th column of $\Zbf$ with $\zbf_k$. 
Let us index the frequencies in the set $\Omega_D$ using a square-bracket, i.e. $\omegabf_{[k]}$ denotes the $k^{th}$ frequency in the set $\Omega_D$.
Define a new matrix $\Zbf_D \in \Cbb^{M \times D}$ with the $i$th row, $k$th column element ${\Zbf_D}_{ i,k}= \frac{1}{\sqrt{N}} e^{-j \xbf_i^T\omegabf_{[k]}}$. Hence, $ \Zbf_D$ is  a sub-matrix of $\Zbf$  formed by only taking the columns corresponding to the frequencies in $\Omega_D$. 
We denote the complement of $\Dset$ with $\Dset^c =\Nset-\Dset$, where $\Nset = \{1,\ldots,N\}$.

\begin{theorem} \label{thm:invertible}
The minimum eigenvalue of  $ \Zbf_D^\herm  \Zbf_D$ satisfies
\begin{align}\label{eqn:thm:invertible}
  \lambda_{min} (  \Zbf_D^\herm  \Zbf_D) \geq \frac{M} {N} ( \lambda_{min}(\Kbf) -t_I),
  \end{align}
  with probability at least $1-\epsilon_I$,   where $t_I \in  (0, \ \lambda_{min}(\Kbf))$ and 
$ \epsilon_I = 2 D \exp( \frac{-t_I^2 M}{2 D\lambda_{min}(\Kbf) (\beta+2/3)})$.
\end{theorem}

\begin{theorem} \label{thm:normsmall}
 Assume that $ \lambda_{min} (  \Zbf_D^\herm  \Zbf_D) \geq \frac{M} {N} ( \lambda_{min}(\Kbf) -t_I)$ for some  $t_I \in  (0, \ \lambda_{min}(\Kbf))$. Then,  we have 
\begin{align}\label{eqn:normsmall}
|| \Zbf_D^\pinv \zbf_l||_2  \leq  \sqrt {D} \frac{t_P +k_{max}}{\lambda_{min}(\Kbf) -t_I} , \quad \forall\, l \in    \Dset^c,
\end{align}
with probability at least $1-\epsilon_P$, where  $\epsilon_P = N^2 \exp (\frac{-t_P^2 M}{14/3})$, $t_P \in (0,2]$. 
\end{theorem}

We now choose $\epsilon_I =\delta/3$, $\epsilon_P =\delta/3$. Using  Thm.~\ref{thm:invertible}, Thm.~\ref{thm:normsmall},  \cite[Prop. 12.15]{foucartRauhut_2013} and re-arranging gives Thm.~\ref{thm:hp:general}.
\iftr 
The proofs of Thm.~\ref{thm:invertible}, Thm.~\ref{thm:normsmall}  and the details of these last steps are provided in   Section~\ref{sec:pf:thm:invertible}, Section~\ref{sec:pf:thm:normsmall}   and Section~\ref{sec:pf:combination},  respectively.  
Section~\ref{sec:pf:lemma:innerproduct} provides the proof of Lemma~\ref{lemma:innerproduct}, which is used in Section~\ref{sec:pf:thm:normsmall}.
\fi

\iftr

\subsection{Proof of Thm.~\protect{\ref{thm:invertible}}} \label{sec:pf:thm:invertible}

Let us define
\begin{align}\label{eqn:zcw}
\zcw_i \triangleq \frac{1}{\sqrt{N}} [e^{+j \xbf_i^T\omegabf_{[1]}}; \ldots; e^{+j \xbf_i^T\omegabf_{[D]}}] \in   \Cbb^{D \times 1}
\end{align}
Hence, $ \Zbf_D$ can be written as  $ \Zbf_D = [\zcw_1^\herm; \ldots;  \zcw_M^\herm]  \in \Cbb^{M \times D}$. 

Let $\bar{\Kbf} \in \Rbb^{ D \times D}$ be the $D\times D$ submatrix of $\Kbf$ that corresponds to the frequencies in $\Omega_D$, i.e. $\bar{K}_{i,j} = k(\omega_{[i]}, \omega_{[j]})$. 
Note that   $\E [\zcw_i \zcw_i^\herm] = \frac{1}{N} \bar{\Kbf} $ due to \eqref{eqn:expectation}. 
We now define $\Vbf_i \in \Cbb^{ D \times D}$ as  
\begin{align} \label{eqn:V:expectation}
\Vbf_i \triangleq \zcw_i \zcw_i^\herm - \E [ \zcw_i \zcw_i^\herm] =  \zcw_i \zcw_i^\herm -  \frac{1}{N} \bar{\Kbf}
\end{align}

Hence,  under statistically independent $\xbf_i$'s, $\Vbf_i$'s are zero-mean,  statistically independent random Hermitian matrices. Let us consider
\begin{align}
\Wbf \triangleq \sum_{i=1}^M \Vbf_i =    \Zbf_D^\herm  \Zbf_D -     \frac{M}{N} \bar{\Kbf}
\end{align}

We will provide bounds on the minimum eigenvalue of $\Zbf_D^\herm  \Zbf_D $  using  bounds on the spectral norm of  $\Wbf$ and the matrix Bernstein inequality:

\vspace{4pt}
 \begin{lemma}[Matrix Bernstein Inequality \cite[Ch.8]{foucartRauhut_2013}]\label{lem:matrixBernstein} 
Let ${{\Vbf}}_1,\ldots,$ ${{\Vbf}}_M \in \mathbb{C}^{D \times D}$  be independent zero-mean Hermitian random matrices. Assume that  $\|{{\Vbf}}_l \| \leq \mu_V$, $\forall l \in \{1,\ldots,M\}$ almost surely. Let
$ 
  \| \sum_{l=1}^M\E[{{\Vbf}}_l^2]  \| \leq  \varrho_V.
$ 
Then, for $t>0$
\begin{align}\label{eqn:matrixBernstein}
 \prob( \| \sum_{l=1}^M {{\Vbf_l}}  \| \geq t ) \leq f_{bn}(\mu_V, \varrho_V,t)
 \end{align}
with  $f_{bn}(\mu_V,\varrho_V,t) \triangleq  2 D \exp\left(-\frac{t^2/2}{\mu_V t/3 + \varrho_V} \right)$.
\end{lemma}
\vspace{4pt}

To bound $\| \Wbf \|$, we first bound  $\| \Vbf_i \|$ as follows: 
\begin{align}
\| \Vbf_i \|  &= \|  \zcw_i \zcw_i^\herm - \frac{1}{N} \bar{\Kbf} \| \\
\label{eqn:mu:trianglein}
                 &\leq   \|  \zcw_i \zcw_i^\herm \| +  \frac{1}{N} \| \bar{\Kbf} \|  \\
 \label{eqn:mu:norm}
                & \leq  2 {D \over N} \triangleq \mu_V
\end{align}
where we have used the triangle inequality  in \eqref{eqn:mu:trianglein}; and $  \|  \zcw_i \zcw_i^\herm \|  = \| \zcw_i \|^2 = {D \over N} $ and  $ \| \bar{\Kbf} \|  \leq \tr({\bar{\Kbf}}) =D$ in \eqref{eqn:mu:norm}.

We now consider $\E[ \Vbf_l^2]$
\begin{align}
\E[ \Vbf_l^2] &=\E [ (\zcw_i \zcw_i^\herm -  \frac{1}{N} \bar{\Kbf})^2] \\
                & =\E [\zcw_i \zcw_i^\herm \zcw_i \zcw_i^\herm -  \frac{1}{N} \zcw_i \zcw_i^\herm \bar{\Kbf} - \frac{1}{N}  \bar{\Kbf} \zcw_i \zcw_i^\herm   \\
                \nonumber        
                 &\quad + \frac{1}{N^2} \bar{\Kbf}^2 ] \\
                &= \frac{D}{N^2}  \bar{\Kbf} -  \frac{1}{N^2}  \bar{\Kbf}^2
\end{align}
where we have used $\zcw_i^\herm \zcw_i = \frac{D}{N}$, and  $\E [\zcw_i \zcw_i^\herm]=  \frac{1}{N} \bar{\Kbf}$. Hence, we have 
\begin{align}
 \sum_{l=1}^M \E[ {\Vbf_l}^2] = M( \frac{D}{N^2}  \bar{\Kbf} -  \frac{1}{N^2}  \bar{\Kbf}^2) \preceq \frac{M \, D}{N^2} \bar{\Kbf}
\end{align}
and 
\begin{align}\label{eqn:var:norm}
\|  \sum_{l=1}^M \E[ {\Vbf_l}^2] \| \leq \frac{M \, D}{N^2} \| \bar{\Kbf} \|  \leq \frac{M \, D}{N^2} \lambda_{max} (\Kbf) \triangleq \varrho_V
\end{align}
where we have used the fact that $ \Abf \preceq \Bbf$ implies $ \| \Abf \| \leq \| \Bbf \| $ and  $\| \bar{\Kbf} \|  \leq  \| \Kbf \| $. 

 Let  $\bar{\epsilon}_I=f_{bn}(\mu_V,\varrho_V,t) =2 D \exp\left(-\frac{t^2/2}{\mu_V t/3 + \varrho_V} \right)$ 
 with  $\mu_V$, $\varrho_V$ from \eqref{eqn:mu:norm}~ and \eqref{eqn:var:norm}, respectively.  
Using Matrix Bernstein Inequality,  with probability  at least $1-\bar{\epsilon}_I$, we have  $ \| \Zbf_D^\herm  \Zbf_D -     \frac{M}{N} \bar{\Kbf} \| < t $ and hence, 
\begin{align}
\lambda_{min}(\Zbf_D^\herm  \Zbf_D)  & \geq  \lambda_{min}(\frac{M} {N} \bar{\Kbf}) - t,\\
 &\geq  \frac{M} {N} \lambda_{min}(\Kbf) - t,\\
  &=   \frac{M} {N} (\lambda_{min}(\Kbf) - t_I),
\end{align}
where we used $\lambda_{min}(\bar{\Kbf}) \geq \lambda_{min}(\Kbf) $,  and we defined the scaled parameter $t_I  =    \frac{N} {M} t$.  
Now we bound $\bar{\epsilon}_I$ as
\begin{align}
\bar{\epsilon}_I &=2 D \exp(- \frac{\frac{1}{2} \frac{M}{N} t_I^2}{2 \frac{D}{N} \frac{1}{3} t_I + \frac{D}{N} \lambda_{max}(\Kbf)}) \\
\label{eqn:deltaI}
 & \leq 2 D  \exp(- \frac{M t_I^2}{2 D (\frac{2}{3} \lambda_{min}(\Kbf)    + \lambda_{max}(\Kbf)}) 
\end{align}
 Note that  the interval of interest for $t_I$  is $t_I \in (0, \lambda_{min}(\Kbf))$. Hence, replacing $t_I$ in the denominator (but not on the numerator)  with  $\lambda_{min}(\Kbf) $ lets us to bound $\bar{\epsilon}_I$ in \eqref{eqn:deltaI}.  Re-arranging and using $\beta= \frac{\lambda_{max}(\Kbf) }{\lambda_{min}(\Kbf)}$ gives the expression in Thm.~\ref{thm:invertible}. 

\subsection{Proof of Thm.~\protect{\ref{thm:normsmall}}} \label{sec:pf:thm:normsmall}
We have  
\begin{align}
\label{eqn:residual:psdinv}
\| \Zbf_D^\pinv \zbf_l \|_2 &= \| (\Zbf_D^\herm \Zbf_D)^{-1} \Zbf_D^\herm \zbf_l\|_2 \\
&\leq  \| (\Zbf_D^\herm \Zbf_D)^{-1} \| \| \Zbf_D^\herm \zbf_l\|_2 \\
&= \frac{1}{\lambda_{min}(\Zbf_D^\herm \Zbf_D)}   \| \Zbf_D^\herm \zbf_l\|_2\\
\label{eqn:residual:bound}
&\leq   \frac{N}{ M( \lambda_{min}(K) -t_I)}  \| \Zbf_D^\herm  \zbf_l\|_2
\end{align}
where  the existence  of $(\Zbf_D^\herm \Zbf_D)^{-1}$ in \eqref{eqn:residual:psdinv} and  \eqref{eqn:residual:bound} follows from the assumption $ \lambda_{min} (  \Zbf_D^\herm  \Zbf_D) \geq \frac{M} {N} ( \lambda_{min}(\Kbf) -t_I)$ for some  $t_I \in  (0, \ \lambda_{min}(\Kbf))$.

We now bound $\| \Zbf_D^\herm \zbf_l\|_2$. Note that columns of $\Zbf_D$ consist of the vectors $\zbf_k$ for $k \in \mathcal{D}$.  
Hence, we have
\begin{align}\label{eqn:zdhzl}
\| \Zbf_D^\herm \zbf_l\|_2 &= \left( \sum_{k \in \mathcal{D}} | \zbf_k^H  \zbf_l |^2  \right)^{1/2}  
\end{align}  
We have the following result that bounds the individual elements in the summation: 
\begin{lemma} \label{lemma:innerproduct}
 Let $k \neq l$.
 We have 
\begin{align}\label{eqn:innerproduct}
|\zbf_k^H  \zbf_l |  \leq  \frac{1}{N} (M k_{max} +t_s)
\end{align}
with probability at least $1\!-\! \epsilon_s$, where  $\epsilon_s \!=\! 2 \exp(-\frac{t_s^2}{2M+ 4t_s/3})$, $t_s \geq 0$.
\end{lemma}

The proof of Lemma~\ref{lemma:innerproduct} is provided in Section~\ref{sec:pf:lemma:innerproduct}.  
Lemma~\ref{lemma:innerproduct} bounds  $|\zbf_k^H  \zbf_l |$ for a given $k,l$ pair: 
$\prob (|\zbf_k^H  \zbf_l  |  \geq   \frac{1}{N} (M k_{max} +t_s)) \leq \epsilon_s$. 
We need a bound that holds for all $k,l$ with $k \neq l$. Hence, we have
\begin{align}
\prob ( \forall k,l  \, |\zbf_k^H  \zbf_l  |  \geq   \frac{1}{N} (M k_{max} +t_s) ) & \leq  \frac{N (N-1)}{2} \epsilon_s \\
\label{eqn:es:N2}
 &\leq   \frac{N^2}{2} \epsilon_s
\end{align}
where we have used the union bound.  Using \eqref{eqn:zdhzl} and \eqref{eqn:es:N2},  with probability at least $1-\frac{N^2}{2} \epsilon_s$ 
\begin{align}
\| \Zbf_D^\herm \zbf_l\|_2 &\leq \left(D  \left(\frac{1}{N} (M k_{max} +t_s)  \right)^2\right)^{1/2} \\
&= \frac{\sqrt{D}}{N} (M k_{max} +t_s) \\
\label{eqn:zdhzl:final} 
&=  \frac{\sqrt{D} M}{N} ( k_{max} +t_P)  
\end{align}
where $t_P= t_s/M$. Hence, rewriting $\epsilon_s$, we have  $\epsilon_s  = 2 \exp(-\frac{ M^2 t_P^2}{2M+ 4M t_P/3}) $.  For $t_P \leq 2$, $\epsilon_s$ can be bounded as 
\begin{align}\label{eqn:es:bound}
\epsilon_s  &\leq 2 \exp(-\frac{ M t_P^2}{2+ 8/3}) = 2 \exp(-\frac{ M t_P^2}{14/3}) 
\end{align}
where we have replaced $t_P$ in the denominator with its upper limit in the interval $t_P \in (0,2]$. Here, we focus on the interval $t_P \in (0,2]$, since this is the interval where \eqref{eqn:ts} with  $t_P= t_s/M$ provides a non-trivial bound. 

Using \eqref{eqn:residual:bound} and \eqref{eqn:zdhzl:final}, we obtain the bound in \eqref{eqn:normsmall} of Thm.~\ref{thm:normsmall}, where the probability expression with $\epsilon_P =\frac{N^2}{2} \epsilon_s$ follows from  \eqref{eqn:es:bound}.

\subsection{Proof of Lemma~\ref{lemma:innerproduct} }\label{sec:pf:lemma:innerproduct}

We note that
\begin{align}
\zbf_k^H  \zbf_l   = \frac{1}{N} \sum_{m=1}^M e^{j (\omegabf_k -\omegabf_l)^T \xbf_m}  =  \frac{1}{N} \sum_{m=1}^M e^{j (\Delta \omegabf_{k,l})^T \xbf_m} 
\end{align}  
where $\Delta \omegabf_{k,l}= \omegabf_k -\omegabf_l$.   
The proof is based on scalar Bernstein inequality \cite[Cor.~7.31]{foucartRauhut_2013}. 
In particular, we define 
\begin{align}
v_m = e^{j (\Delta \omegabf_{k,l})^T \xbf_m}  - \E [e^{j (\Delta \omegabf_{k,l})^T \xbf_m} ] \\
       = e^{j (\Delta \omegabf_{k,l})^T \xbf_m}  - k(\Delta \omegabf_{k,l})
\end{align} 
We note  that $\E[v_m] = 0$ and $v_m$ are independent random variables.   We have
\begin{align}
  | v_m | \leq |e^{j (\Delta \omegabf_{k,l})^T \xbf_m}|  +| k(\Delta \omegabf_{k,l}) | \leq 2
\end{align}
and
\begin{align}
 \E [ | v_m |^2 ]
  &=\E[ | e^{j (\Delta \omegabf_{k,l})^T \xbf_m}  - k(\Delta \omegabf_{k,l})|^2] \\
  \nonumber
  &= \E[ 1 - k(\Delta \omegabf_{k,l}) (e^{j (\Delta \omegabf_{k,l})^T \xbf_m} + e^{-j (\Delta \omegabf_{k,l})^T \xbf_m} ) \\ &+ k^2(\Delta \omegabf_{k,l})  ] \\
  \label{eqn:vm2}
   &= 1 -   k^2(\Delta \omegabf_{k,l}) 
\end{align}
where we have used \eqref{eqn:expectation} in \eqref{eqn:vm2}.  
Hence, we have 
\begin{align}
 \sum_{m=1}^M  \E [ | v_m |^2 ] = M (1 -   k^2(\Delta \omegabf_{k,l})) \leq M 
 \end{align}
Now, by scalar Bernstein inequality \cite[Cor.~7.31]{foucartRauhut_2013}, with  probability at least $1-\epsilon_s$ with  $\epsilon_s\! =\! 2 \exp(-\frac{t_s^2}{2M+ 4t_s/3})$, we have
\begin{align}\label{eqn:ts}
|  \sum_{m=1}^M e^{j (\Delta \omegabf_{k,l})^T \xbf_m}  - M k(\Delta \omegabf_{k,l})|  < t_s
\end{align}
where $t_s \geq 0$. In particular, note that \eqref{eqn:ts} provides a non-trivial bound for $t_s \leq 2 M$. 
Eqn.~\eqref{eqn:ts} implies
\begin{align}
 | \frac{1}{N} \sum_{m=1}^M  e^{j (\Delta \omegabf_{k,l})^T \xbf_m}  |  
& <   \frac{1}{N}  ( M | k(\Delta \omegabf_{k,l})| +    t_s) \\
& \leq   \frac{1}{N} ( M k_{max} +  t_s)
\end{align}
which is the desired inequality in \eqref{eqn:innerproduct}.

\subsection{Combining Thm.~\protect{\ref{thm:invertible}}  and Thm.~\protect{\ref{thm:normsmall}}} \label{sec:pf:combination}

We define the following events: 
\begin{align}
\Einj  &\triangleq \{\ \Zbf_D \text{ is injective} \}\\ 
\Enorm(\eta) &\triangleq \{ \| \Zbf_D^\pinv \zbf_l \|_2 \leq \eta,\,\, \forall\, l \in    \Dset^c, \eta>0 \}\\
\EBP &\triangleq  \{ \text{The unique minimizer of \eqref{eqn:bp:z} is $\bar{\thetabf}$} \}
\end{align}
Note that with $M \geq D$, $\Einj $ states that $\rank (\Zbf_D )=D$, i.e.   $\lambda_{min} (  \Zbf_D^\herm  \Zbf_D)>0$. 
We use the following result:

\begin{lemma} \cite[Prop. 12.15]{foucartRauhut_2013}]\label{lem:norm2BP} 
Let $\bar{\thetabf} \in \Cbb^{N \times 1}$ be a D-sparse vector such that  $\sgn(\bar{\thetabf}_D)$ forms a Rademacher or Steinhaus sequence. Let $\epsilon_\eta= 2 N \exp(-\eta^{-2}/2)$. Assume that $\Zbf \in \Cbb^{M \times N}$ is such that  $\Einj$  and $\Enorm(\eta)$ hold.  Then,  with probability at least $1-\epsilon_\eta$,  $\EBP$ holds. 
\end{lemma}

Equivalently, Lemma~\ref{lem:norm2BP} states $\prob(\EBP^c | \Einj,  \Enorm(\eta)) \leq \epsilon_\eta$, where $\EBP^c$ is the complement of $\EBP$.  Similarly, Thm.~{\ref{thm:invertible}}  implies   $\prob( \Einj^c) \leq \epsilon_I$   and  Thm.~\ref{thm:normsmall} provides bounds on   $\prob( \Enorm^c(\eta) | \Einj) $ with an appropriate choice of constants. 
Let $ \Einjnorm= \{  \Einj  \text{ and }  \Enorm(\eta) \}$. We consider the following bound on the probability that basis pursuit fails
\begin{align}
\prob(\EBP^c ) &\!=\! \prob(\EBP^c | \Einjnorm ) \prob(\Einjnorm) \!+\! \prob(\EBP^c | \Einjnorm^c ) \prob(\Einjnorm^c) \\
& \leq  \prob(\EBP^c | \Einjnorm )  + \prob(\Einjnorm^c) \\
\nonumber
& = \prob(\EBP^c | \Einjnorm )  + \prob( \Einj^c U \Enorm^c(\eta) |\Einj^c ) \prob(\Einj^c ) \\
&\quad + \prob( \Einj^c U \Enorm^c(\eta) |\Einj ) \prob(\Einj ) \\
\label{eqn:ebpc:bound}
& \leq   \prob(\EBP^c | \Einjnorm )  + \prob( \Einj^c) + \prob( \Enorm^c(\eta) | \Einj) 
\end{align}

We will now consider the events in \eqref{eqn:ebpc:bound} one by one  in order to provide sufficient conditions in terms of the number of data points  $M$ so that  $\EBP$ holds. 

Let $\prob( \Einj^c) \leq \epsilon_I =\delta/3$, where $\delta \in [0,1]$.  Then, by Thm.~\ref{thm:invertible},  the following condition   on  $M$  guarantees that $\prob( \Einj^c) \leq \delta/3$ 
\begin{align}\label{eqn:M:invertible}
M \geq \frac{1}{t_I^2} C_{\beta} \, D\, \ln(6 {D \over \delta})
\end{align}
where $ C_{\beta} =2  (\beta+\frac{2}{3}) \lambda_{min}(\Kbf)$. 

Let $\prob(\Enorm^c(\eta) | \Einj) \leq \epsilon_P=\delta/3$, $\eta= \sqrt {D} \frac{t_P +k_{max}}{\lambda_{min}(\Kbf) -t_I}$.  Then, by Thm.~\ref{thm:normsmall},  the following condition  on  $M$ guarantees     that $\prob(\Enorm^c(\eta) | \Einj) \leq \delta/3$, 
\begin{align}\label{eqn:M:norm}
M &\geq \frac{1}{t_P^2}  {14\over 3} \ln(3 {N^2 \over \delta}) 
\end{align}
which is implied by 
\begin{align}
M & \geq  \frac{1}{t_P^2}  {28 \over 3} \ln(3 {N \over \delta}), 
\end{align}
where we have used $\ln(3 {N^2 \ \delta}) = \ln(3 {N / \delta}) + \ln(N) \leq  2 \ln(3 {N / \delta}) $  for $\delta \in [0,1]$.  
Now set $t_P= { t_I \over \sqrt{D}} C_\eta$ with $ C_\eta  = \sqrt{\frac{28}{3} \frac{1}{C_\beta}}$. 
Hence, with $2 D \leq N$,  the following condition guarantees \eqref{eqn:M:invertible} 
\begin{align}
\label{eqn:M:norm:rev}
 M &\geq   \frac{1}{t_I^2} C_{\beta} \, D\,  \ln(3 {N \over \delta}) 
\end{align}
Hence, \eqref{eqn:M:norm:rev} implies $\prob(\Enorm^c(\eta) | \Einj) \leq \delta/3$ and $\prob( \Einj^c) \leq \delta/3$.

Let $\prob(\EBP^c | \Einjnorm ) =\prob(\EBP^c | \Einj,  \Enorm(\eta)) \leq \epsilon_\eta = \delta/3$. Then, by Lemma~\ref{lem:norm2BP}, the following condition on $t_I$ guarantees  that  $\prob(\EBP^c | \Einj,  \Enorm(\eta)) \leq  \delta/3$, 
\begin{align}\label{eqn:tI:eta}
\frac{1}{t_I^2} \geq C_q,
\end{align}
where $C_{q} = (\frac{1+q C_{\eta}}{\lambda_{min} - q  \sqrt{D} k_{max}})^2 $ and $q=\sqrt{2\ln (6 N/\delta)}$.  
Using \eqref{eqn:tI:eta}, \eqref{eqn:M:norm:rev} can be rewritten as
\begin{align}\label{eqn:M:norm:final}
M \geq  C_{q}  C_{\beta} \, D\,  \ln(3 {N \over \delta}) 
\end{align}
Hence,   \eqref{eqn:M:norm:final} is a sufficient condition for bounding
each term in the right-hand side of \eqref{eqn:ebpc:bound} with $\delta/3$.  Hence, if the number of data points satisfy \eqref{eqn:M:norm:final}, we have $ \prob(\EBP^c )  \leq \delta$, which is the desired condition in \eqref{eqn:M} in Thm.~\ref{thm:hp:general}. Note that the condition $t_I \in  (0, \ \lambda_{min}(\Kbf))$ of  Thm~\ref{thm:invertible} is satisfied under \eqref{eqn:tI:eta} and the condition $\lambda_{min}(\Kbf) \geq q  \sqrt{D} k_{max}$. Similarly, it can be shown that  $t_P \in (0,2]$ of Thm~\ref{thm:normsmall} is satisfied under $t_P= { t_I \over \sqrt{D}} C_\eta$, $ C_{\eta}/\sqrt{C_{q}}  \leq 2 \sqrt{D} $ using straightforward algebraic substitutions.

\fi

\newpage
\bibliographystyle{ieeetr}
\bibliography{\bibdirM/JNabrv,\bibdirM/bib_ayca,\bibdirMM/bib_randomMatrices,\bibdirM/bib_distributedEstimation,\bibdirM/bib_compressiveSampling2,\bibdirMM/bib_eh_M2M,\bibdirM/bib_energyHarvesting,\bibdirM/bib_csecuritySignal,\bibdirM/bib_robust,\bibdirMM/bib_robust_eh,\bibdirM/bib_optimization,\bibdirM/bib_books,\bibdirM/bib_linearEncoding,\bibdirM/bib_GPKernelCSAdaptive,\bibdirM/bib_RFF}

\end{document}